\pdfoutput=1

\documentclass[11pt]{article}

\usepackage{acl}

\usepackage{times}
\usepackage{latexsym}

\usepackage[T1]{fontenc}

\usepackage[utf8]{inputenc}

\usepackage{microtype}

\usepackage{inconsolata}
\usepackage{algorithm}
\usepackage[noend]{algpseudocode}
\usepackage{booktabs}
\usepackage{amsfonts}
\usepackage{amsmath}
\usepackage{bm}
\usepackage{graphicx}
\usepackage{subfig}
\usepackage{multirow}
\usepackage{array}
\usepackage{colortbl}
\usepackage{tabularx}

\title{RadialRouter: Structured Representation for Efficient and Robust Large Language Models Routing}

\author{
    Ruihan Jin\textsuperscript{\rm 1} \quad 
    Pengpeng Shao\textsuperscript{\rm 1}\thanks{Corresponding authors.}\quad 
    Zhengqi Wen\textsuperscript{\rm 1}\footnotemark[1]\quad 
    Jinyang Wu\textsuperscript{\rm 1} \\ \bf 
    Mingkuan Feng\textsuperscript{\rm 1} \quad 
    Shuai Zhang\textsuperscript{\rm 1} \quad 
    Jianhua Tao\textsuperscript{\rm 1,2} \\
    \textsuperscript{\rm 1}Department of Automation, Tsinghua
University\\
  \textsuperscript{\rm 2}Beijing National Research Center for
Information Science and Technology\\
    \texttt{jinrh24@mails.tsinghua.edu.cn} \qquad \texttt{\{ppshao, zqwen\}@tsinghua.edu.cn}
}

\begin{document}
\maketitle
\begin{abstract}
The rapid advancements in large language models (LLMs) have led to the emergence of routing techniques, which aim to efficiently select the optimal LLM from diverse candidates to tackle specific tasks, optimizing performance while reducing costs. Current LLM routing methods are limited in effectiveness due to insufficient exploration of the intrinsic connection between user queries and the characteristics of LLMs. To address this issue, in this paper, we present \textbf{RadialRouter}, a novel framework for LLM routing which employs a lightweight Transformer-based backbone with a radial structure named \textbf{RadialFormer} to articulate the query-LLMs relationship. The optimal LLM selection is performed based on the final states of RadialFormer. The pipeline is further refined by an objective function that combines Kullback-Leibler divergence with the query-query contrastive loss to enhance robustness. Experimental results on RouterBench show that RadialRouter significantly outperforms existing routing methods by 9.2\% and 5.8\% in the \textit{Balance} and \textit{Cost First} scenarios, respectively. Additionally, its adaptability toward different performance-cost trade-offs and the dynamic LLM pool demonstrates practical application potential.
\end{abstract}

\begin{figure}[t]
  \centering
  \includegraphics[width=0.48\textwidth]{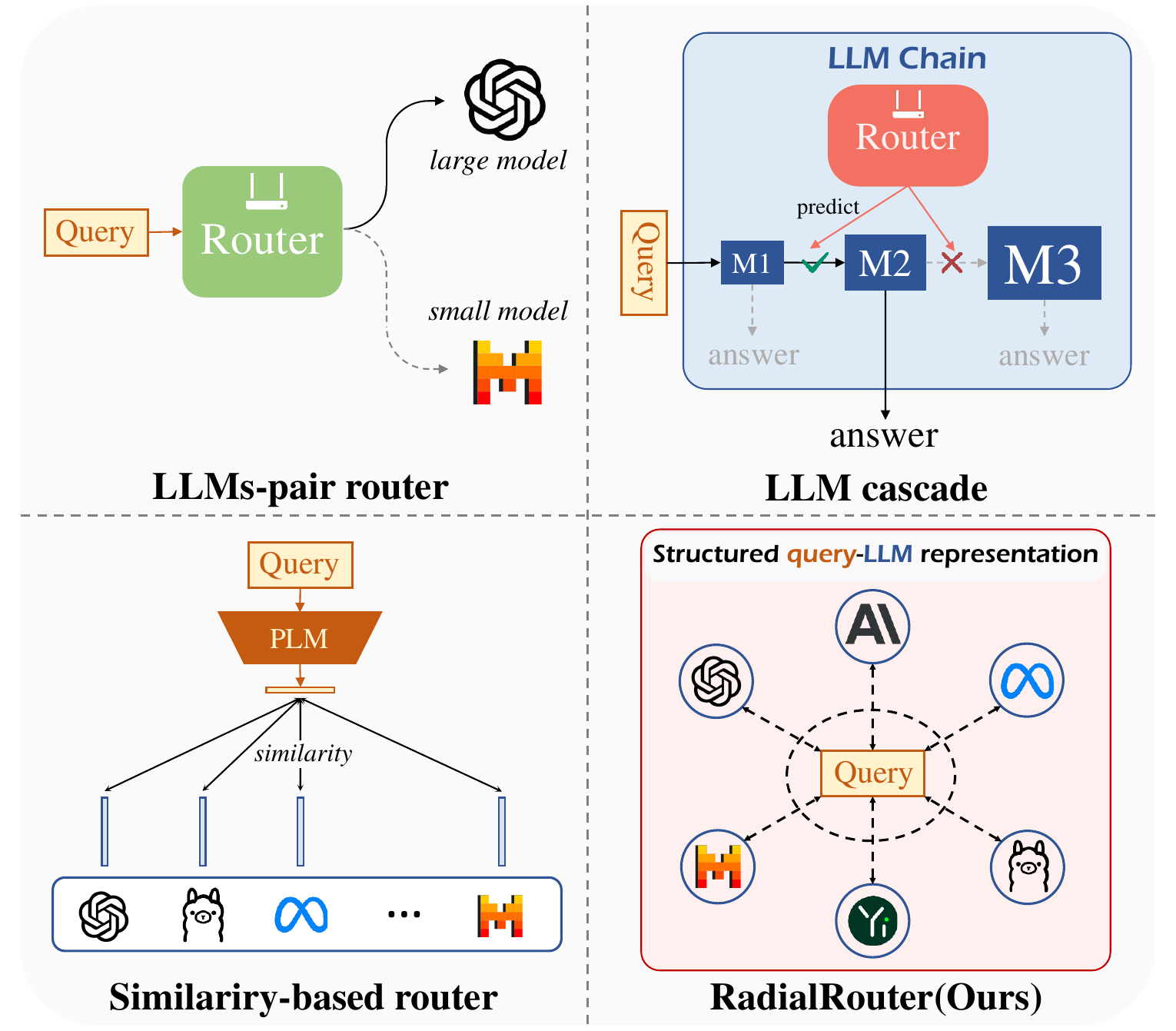}
  \caption{\textbf{Paradigm comparison between different LLM routing methods.} Existing methods lack the modeling of the interrelation between the query and LLMs, while the proposed RadialRouter unifies the routing process in a structured representation.}
  \label{fig:introduction}
  \vspace{-0.5cm}
\end{figure}

\section{Introduction}

Recent advances in natural language processing, significantly driven by the development of large language models (LLMs), has opened new frontiers across numerous applications. LLMs demonstrate outstanding performance on various tasks, including mathematical problem-solving \cite{romera2024mathematical}, commonsense reasoning \cite{zhao2023large}, and code generation \cite{wu2023autogen}. The growing reliance on LLMs gives rise to the concept of \textit{LLM ensemble} \cite{chen2025harnessing}, which integrates multiple LLMs to establish a system capable of multitasking, thereby generating more accurate and robust responses to user inputs. Through this collaborative approach, practitioners can leverage the unique strengths of different LLMs, potentially improving the overall performance and reliability in addressing diverse requirements. However, as the size and complexity of LLMs increase, the challenges of deploying LLM ensemble \textemdash such as computational cost, latency, and scalability \textemdash also intensify.

To address these challenges, \textit{LLM routing} is explored to dynamically assign a specific LLM in the LLM ensemble to user queries. As shown in Fig.\ref{fig:introduction}, early attempts \cite{ding2024hybrid, ong2024routellm} employ a binary score router to decide whether to select a smaller LLM or a larger one for a given query. \cite{chen2023frugalgpt} utilizes a router to guide cascaded LLMs for generating responses. Further research \cite{chen2024routerdc} introduces similarity matching to align the query with LLMs and select the most appropriate LLM. However, these methods are limited in the following aspects: 1) They narrow the routing sorely to identify the optimal LLM, failing to capture the intrinsic connection between the query and LLMs, which undermines their effectiveness in achieving ideal routing results. 2) The exclusive dependence on the features extracted by the BERT-based text encoder hinders their ability to leverage the contextual information and prevents a nuanced understanding of the underlying relationships and constraining the efficiency of routing. 3) Existing methods that limit routing to a fixed number of LLMs struggle to adapt to a dynamically evolving pool of LLMs. 4) Certain approaches neglect the actual requirements of the task, rendering them ineffective in scenarios that necessitate a simultaneous consideration of both performance and cost.

In this paper, we propose a novel LLM routing approach named \textbf{RadialRouter}, which leverages a Transformer-based architecture to enhance performance for efficient and robust LLM routing. To represent the interrelationship between the query and LLMs, we propose \textbf{RadialFormer} as the backbone of RadialRouter and incorporate the structure consisting of a relay node and $n$ satellite nodes. During update, these nodes are processed through the multi-head attention mechanism. Compared with the standard Transformer, RadialFormer reduces the computational complexity from $O(l^2d)$ to $O(ld)$ given the sequence length $l$ and the dimension of the hidden state $d$. The optimal LLM selection is performed based on the final states of satellite nodes. To guide the selection and enhance comprehensive representation, we employ a Kullback-Leibler divergence loss for supervision. Additionally, we cluster the queries into groups and introduce a query-query contrastive loss, which fosters the generation of similar embeddings for semantically related queries, facilitating robust LLM routing.

We conduct experiments on challenging RouterBench \cite{hu2024routerbench} encompassing 4 task domains (commonsense reasoning, knowledge-based language understanding, math, and coding) to evaluate the proposed RadialRouter in three scenarios. Extensive experiments demonstrate that RadialRouter efficiently harnesses the interrelationship of routing and outperforms existing routing methods by a large margin. Furthermore, the adaptability of RadialRouter toward different performance-cost trade-offs and the dynamic LLM pool is verified through extended experiments.

Our contributions are summarized as follows:

\begin{itemize}
    \vspace{-0.2cm}
    \setlength{\itemsep}{0cm}
    \item We propose RadialRouter, a novel framework that leverages a Transformer-based architecture to dynamically route user queries to suitable LLMs.
    \item We introduce a lightweight architecture RadialFormer as the backbone of RadialRouter to capture the interrelationship between the query and LLMs in routing. To improve the robustness of routing, we incorporate contrastive loss in the optimization of RadialRouter.
    \item Experimental results show that RadialRouter outperforms baseline routing methods and achieves efficient and robust routing in three scenarios with different performance-cost trade-offs.
\end{itemize}

\begin{figure*}[t]
  \centering
  \subfloat[The framework of our proposed \textbf{RadialRouter}. RadialRouter captures the interrelation between the query and LLMs through a lightweight Transformer-based architecture. It selects the optimal LLM based on the final states of satellite nodes, while optimizing the pipeline through an objective function that combines Kullback-Leibler divergence and query-query contrastive loss.]{
  \includegraphics[width=0.66\textwidth]{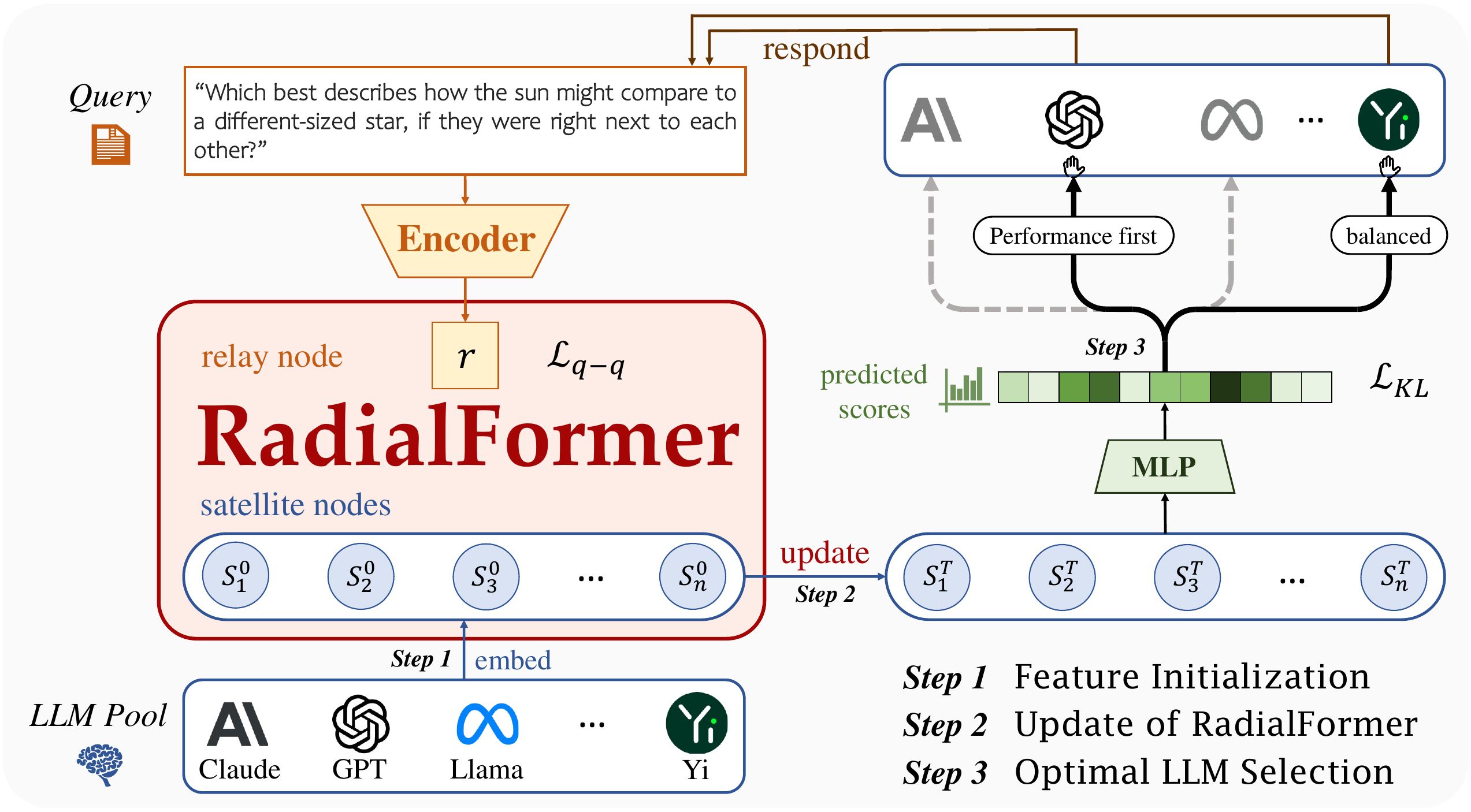}
  \label{fig:radialrouter}}
  \hfill 
  \subfloat[Connections of one layer in \textbf{RadialFormer}, where each satellite node is exclusively connected to the relay node to form a radial configuration.]{
  \includegraphics[width=0.29\textwidth]{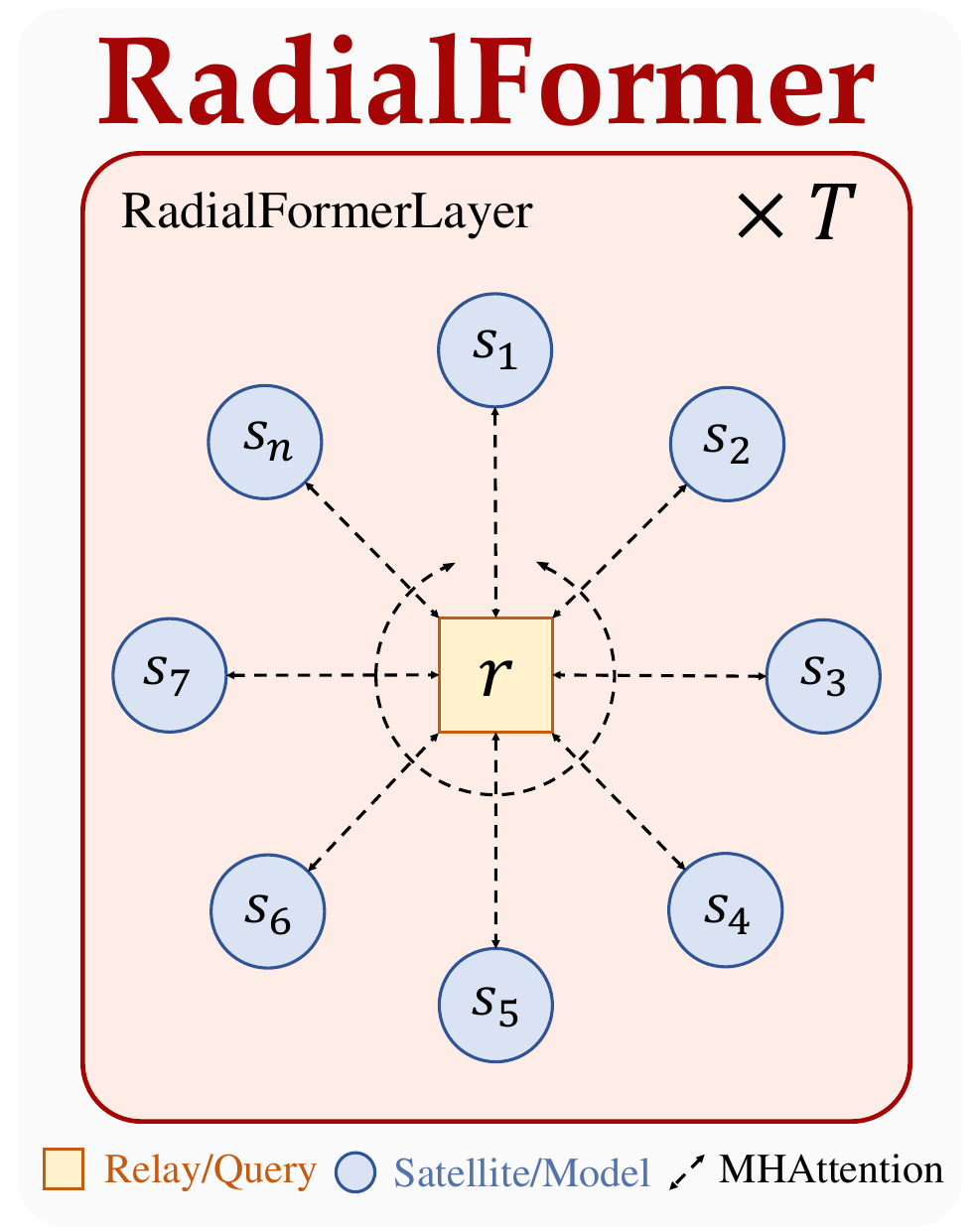}
  \label{fig:radialformer}}
  \caption{Overview of \textbf{RadialRouter} methodology.}
  \label{fig:method}
\end{figure*}

\section{Related Work}

\paragraph{LLM Ensemble}

The remarkable performance exhibited by a single LLM, coupled with the increasing demand for enhanced cross-domain capabilities, has catalyzed the emergence of the concept of LLM ensemble. Majority voting \cite{wang2022self, li2024more} is a simple yet effective method to achieve the LLM ensemble. \cite{jiang2023llm} proposes a supervised ensembling method to produce enhanced results by synthesizing the outputs of all LLMs. \cite{du2023improving} develops a debate framework for LLM collaboration. LLM cascading \cite{chen2023frugalgpt, yue2023large, gupta2024language, nie2024online} adopts a serialized model architecture and halts when the output quality is satisfied. \cite{wang2023fusing} tackles the fusion-of-experts problem by combining outputs from models with diverse knowledge domains. These approaches frequently exhibit considerable computational complexity, leading to substantial time latency and cost in practical applications. In contrast, our RadialRouter is highly efficient, as it requires only a single invocation of the routed LLM.

\paragraph{LLM Routing}

Similar to Mixture-of-Expert (MoE) approaches \cite{jacobs1991adaptive, collobert2003scaling, jiang2024mixtral}, LLM routing is designed to identify the most suitable LLM for a query, allowing the lightweight activation of LLM ensemble. \cite{shnitzer2023large} addresses the LLM selection problem with a series of binary classification tasks. ZOOTER \cite{lu2023routing} develops a reward-guided method to train the routing function. HybridLLM \cite{ding2024hybrid} proposes a hybrid approach to save cost and maintain quality leveraging LLM pairs. The framework proposed in RouteLLM \cite{ong2024routellm} dynamically routes between a strong model and a weak model. RouterDC \cite{chen2024routerdc} improves the routing performance by introducing dual contrastive learning. Recent advancements in LLM routing utilizing graph neural networks \cite{feng2024graphrouter} and reinforcement learning \cite{sikeridis2024pickllm, yue2025masrouter} show significant promise. Different from the aforementioned methods, the proposed RadialRouter leverages a Transformer-based backbone to achieve performance-cost balanced LLM routing.

\section{Method}


We consider a set of candidate LLMs $\{\mathrm{LLM}_i: i=1,\dots n\}$, which can include local open-source LLMs and off-the-shelf LLMs hosted on cloud platforms. Our goal is to learn a router to select the most suitable LLM for each user query. For each round, the router receives the user query $\mathrm{x}$ as input and chooses the optimal $\mathrm{LLM}_{\hat{i}}$ for response, balancing high performance with minimal cost.

In this section, we propose \textbf{RadialRouter}, a framework that leverages a Transformer-based architecture for query-based LLM routing. Fig.\ref{fig:radialrouter} shows the framework of the proposed method. We present \textbf{RadialFormer}, a lightweight Transformer-based architecture as the backbone of routing (Sec.\ref{sec:radialformer_architecture}). Based on the update of RadialFormer, we perform optimal LLM selection (Sec.\ref{sec:optimal_llm_selection}) and introduce a contrastive loss to optimize the router (Sec.\ref{sec:optimization_with_contrastive_loss}). The overall algorithm of RadialRouter is illustrated in Alg.\ref{alg:radialrouter} of Appendix \ref{sec:training_and_inference_algorithm_of_radialrouter}.

\subsection{RadialFormer Architecture}
\label{sec:radialformer_architecture}

The key to LLM routing lies in designing a scoring mechanism to measure the potential capacity of LLMs on user queries. The backbone of the router necessitates the efficient representation of the query and LLMs. In this work, we present a novel Transformer-based architecture named RadialFormer, which builds upon the foundational design of the Star-Transformer \cite{guo2019star}, incorporating specific enhancements tailored to LLM routing tasks.

The RadialFormer consists of one relay node and $n$ satellite nodes, representing the input user query and $n$ candidate LLMs, respectively. The topology of the model is simplified for computational efficiency, where each satellite node is exclusively connected to the relay node, forming a radial configuration as shown in Fig.\ref{fig:radialformer}. Through the computational mechanisms employed by RadialFormer, the interrelationship between the query and LLMs is comprehensively captured, providing valuable insights for effective routing.

\def\br{\mathbf{r}}
\def\bs{\mathbf{s}}

\paragraph{Update of RadialFormer} Let $\br^t\in\mathbb{R}^{1\times d}$ denotes the state of the relay node, and $\mathbf{S}^t\in\mathbb{R}^{n\times d}$ denote the states of the $n$ satellite nodes at time step $t$. Given a query $\mathrm{x}$, we initialize the relay node with the query embedding encoded by a pre-trained language model as $\mathbf{q}=\mathcal{E}(\mathrm{x})\in\mathbb{R}^{1\times d}$. The satellite nodes are initialized with $n$ learnable model embeddings $\{\mathbf{m}_i:i=1,\dots,n\}$.

Based on the multi-head attention mechanism \cite{vaswani2017attention}, the update of RadialFormer focuses on processing the relay node and the satellite nodes. Details of the multi-head attention mechanism are introduced in Appendix \ref{sec:multi_head_attention}. The satellite node $\bs_i$ is updated from the contextual information considering the relay node $\br^{t-1}$, the previous state $\bs_i^{t-1}$, and the initial state $\mathbf{m}_i$:
\begin{equation}
\mathbf{C}^t_i = [\bs^{t-1}_{i}; \mathbf{m}_i; \br^{t-1}],
\end{equation}
\begin{equation}
\bs^t_i = \mathrm{MHAttn}(\bs^{t-1}_i, \mathbf{C}^t_i),
\end{equation}
where $\mathbf{C}^t_i$ denotes the contextual information of the $i$-th satellite node. The relay node $\br$ is updated from all the satellite nodes and its previous state:
\begin{equation}
\br^t = \mathrm{MHAttn}(\br^{t-1}, [\br^{t-1};\mathbf{S}^t]).
\end{equation}

The updated satellite nodes are regularized through layer normalization \cite{ba2016layer}. The update algorithm of RadialFormer is shown in Alg.\ref{alg:radialformer}. Given the sequence length $l$ and the dimension of the hidden state $d$, RadialFormer reduces the computational complexity of the standard Transformer from $O(l^2d)$ to $O(ld)$. The specific design of RadialFormer integrates both lightweight and structured representations of the routing. Subsequent experiments validate that RadialFormer facilitates efficient and effective LLM routing.

\subsection{Optimal LLM Selection}
\label{sec:optimal_llm_selection}

Following the update of RadialFormer, we sent the final states $\mathbf{S}^T$ to an MLP network $\mathcal{M}$ to predict the potential score of the corresponding LLM as $\bs_i^{\mathrm{pr}}=\mathcal{M}(\bs_i^T)$, which comprehensively integrates both the information of the query and LLMs through RadialFormer. The optimal LLM selection is performed based on the predicted scores as $\hat{i}=\arg\max_i(p_i)$, where $p_i=\mathrm{softmax}(\bs_i^{\mathrm{pr}})$ denotes the routing probability.

Given a query $\mathrm{x}_j$, the process of LLM selection is supervised by the scores of the candidate LLMs $\mathrm{score}_{\mathrm{x}_j}=\{s_j^{(i)}:i=1,\dots,n\}$, which we identify in advance (described in Sec.\ref{sec:metrics}). We employ the Kullback-Leibler divergence loss \cite{kullback1951information} for supervision to guide the routing probability toward the probability derived from the exponent of true scores, which is defined as:
\begin{equation}
    \mathcal{L}_{\text{KL}}(\mathrm{x};\bm{\theta})=D_{\mathrm{KL}}(p\|q)=\sum_{i=0}^n p_i\log\frac{p_i}{q_i},
    \label{eq:kullback-leibler}
\end{equation}
where $\bm{\theta}$ denotes the parameters in RadialRouter, $p$ and $q=\mathrm{softmax}(\mathrm{score}_x)$ denote the predicted routing probability and the ground truth probability, respectively.

\begin{algorithm}[t]
    \caption{Update of RadialFormer}
    \label{alg:radialformer}
    \textbf{Input}: number of layers $T$, model embeddings $\mathbf{m}_1, \ldots, \mathbf{m}_n$, and query embedding $\mathbf{q}$.
    \begin{algorithmic}[1]
    \State \textcolor[rgb]{0.00, 0.59, 0.00}{{// \textit{Feature Initialization}}}
    \State $\bs^0_1,\ldots,\bs^0_n\leftarrow\mathbf{m}_1,\ldots,\mathbf{m}_n$
    \State $\br^0 \leftarrow \mathbf{q}$
    \For{$t=1$ \textbf{to} $T$}
        \State \textcolor[rgb]{0.00,0.59,0.00}{{// \textit{Update the satellite nodes}}}
        \For{$i=1$ \textbf{to} $n$}
            \State $\mathbf{C}^t_i \leftarrow [\bs^{t-1}_{i}; \mathbf{m}_i; \br^{t-1}]$
            \State $\bs^t_i \leftarrow \mathrm{MHAttn}(\bs^{t-1}_i, \mathbf{C}^t_i)$
            \State  $\bs^t_i \leftarrow \mathrm{LayerNorm}(\mathrm{ReLU}(\bs^t_i))$
        \EndFor
        \State \textcolor[rgb]{0.00,0.59,0.00}{{// \textit{Update the relay node}}}
        \State $\br^t \leftarrow \mathrm{MHAttn}(\br^{t-1}, [\br^{t-1};\mathbf{S}^t])$
        \State $\br^t \leftarrow \mathrm{LayerNorm}(\mathrm{ReLU}(\br^t))$
    \EndFor
    \end{algorithmic}
\end{algorithm}

\subsection{Optimization with Contrastive Loss}
\label{sec:optimization_with_contrastive_loss}

Inspired by \cite{chen2024routerdc}, we leverage a contrastive loss to provide additional supervision for the optimization of RadialFormer.

\paragraph{Query-Query Contrastive Loss}

To enhance the robustness of LLM routing, we introduce a query-query contrastive loss, which promotes the ability of the language encoder in RadialRouter to generate analogous embeddings for semantically similar queries. Following \cite{chen2024routerdc}, we transform the query embeddings encoder by a pre-trained language model into low-dimensional vectors by the t-SNE algorithm \cite{van2008visualizing} and perform $k$-means clustering \cite{macqueen1967some} to obtain $N$ semantic groups $\{\mathcal{K}_1,\ldots,\mathcal{K}_N\}$. We use the sample-sample contrastive loss to promote the generation of embeddings, formulated as:
\begin{equation}
    \begin{aligned}
    &\mathcal{L}_{\text{q-q}}(\mathrm{x};\bm{\theta})= \\
    &-\log\frac{e^{\mathrm{sim}\langle\mathcal{E}(\mathrm{x}),\mathcal{E}(\mathrm{x}^+)\rangle}}{e^{\mathrm{sim}\langle\mathcal{E}(\mathrm{x}),\mathcal{E}(\mathrm{x}^+)\rangle}+\sum\limits_te^{\mathrm{sim}\langle\mathcal{E}(\mathrm{x}),\mathcal{E}(\mathrm{x}_t^-)\rangle}},
    \label{eq:query_query}
    \end{aligned}
\end{equation}
where $\mathrm{x}^+$ denotes in-group query, $\mathrm{x}_t^-$ denotes out-group queries, and $\mathrm{sim}\langle\cdot,\cdot\rangle$ denotes the cosine similarity.

\paragraph{Optimization Objective}

Finally, we learn the RadialRouter by minimizing a final objective that combines the KL divergence and the query-query contrastive loss as:
\begin{equation}
    \bm{\theta}^*=\arg\min\underset{{\mathrm{x}\sim\mathcal{D}_{\mathrm{train}}}}{\mathbb{E}}\mathcal{L}_{\text{KL}}(\mathrm{x};\bm{\theta})+\lambda\mathcal{L}_{\text{q-q}}(\mathrm{x};\bm{\theta}),
    \label{eq:optimization}
\end{equation}
where $\lambda>0$ is a hyper-parameter.

\section{Experimental Setup}

\subsection{Datasets and Candidate LLMs}

We conduct experiments on RouterBench\cite{hu2024routerbench} to compare our RadialRouter model with baselines considering both performance and costs. We select user queries from 6 representative datasets in RouterBench across 4 task domains: (i) \textbf{Commonsense Reasoning}: Hellaswag \cite{zellers2019hellaswag}, Winogrande \cite{sakaguchi2021winogrande}, ARC Challenge \cite{clark2018think}; (ii) \textbf{Knowledge-based Language Understanding}: MMLU \cite{hendryckstest2021}; (iii) \textbf{Math}: GSM8K \cite{cobbe2021training}; (iv) \textbf{Coding}: MBPP \cite{austin2021program}. A total of 11 candidate LLMs are involved in RouterBench, including both \textbf{open-source models}: Llama-70B-chat \cite{touvron2023llama}, Mixtral-8x7B-chat \cite{aggarwal2024automix}, Yi-34B-chat \cite{young2024yi}, Code Llama-34B \cite{roziere2023code}, Mistral-7B-chat \cite{jiang2023mistral}, WizardLM-13B \cite{xu2024wizardlm}; and \textbf{proprietary models}: GPT-4, GPT-3.5-turbo \cite{achiam2023gpt}, Claude-instant-v1, Claude-v1, Claude-v2 \cite{anthropic2023model}.

\subsection{Baselines}

RadialRouter is compared with the following routing methods: (i) \textbf{CosineClassifier} trains a cosine classifier on the query embedding and performs a multi-class classification on candidate LLMs, which can be regarded as a simplified version of \cite{chen2024routerdc}. (ii) \textbf{HybridLLM} \cite{ding2024hybrid} trains a language model to categorize queries to either small or large LLM. Mistral-7B-chat and GPT-4 are chosen as the small are large LLM, as they have the highest and lowest cost, respectively. We use DeBERTa \cite{he2020deberta} as the router model. (iii) \textbf{FrugalGPT} \cite{chen2023frugalgpt} uses a pre-trained language model to learn the scores of the generated results and guide the LLM cascade. We also use DeBERTa as the prediction model. (iv) \textbf{RouterDC} \cite{chen2024routerdc} learns a router to select the suitable LLM for user queries by dual contrastive learning. (v) \textbf{GraphRouter} \cite{feng2024graphrouter} introduces a graph-based framework to leverage contextual information among tasks, queries, and LLMs for routing.

\subsection{Metrics}
\label{sec:metrics}

Metrics that consider both performance and cost are utilized to evaluate RadialRouter and baselines.

\begin{itemize}
    \setlength{\itemsep}{0cm}
    \item \textbf{Performance} refers to the average accuracy of responses across user queries generated by LLM or LLM ensemble equipped with routing methods.
    \item \textbf{Cost} refers to the average LLM inference cost for generating responses to the queries, which is expressed in dollars. The statistics of candidate LLMs on RouterBench are shown in Appendix \ref{sec:statistics_of_candidate_llms}.
    \item \textbf{Score} is employed to assess how effectively a method balances performance and cost: for an input query $\mathrm{x}_j$, the score for $\mathrm{LLM}_i$ is calculated via
    \begin{equation}
    \mathrm{score}_{ij}=\mathrm{performance}_{ij}-\alpha\cdot \mathrm{cost}_i,
    \label{eq:score}
    \end{equation}
    where $\alpha$ balances the performance-cost trade-off and a higher $\alpha$ indicates a preference for saving cost. We define three scenarios: \textit{Performance First}, \textit{Balance}, and \textit{Cost First}, which correspond to different priorities between performance and cost. In three scenarios, we set the value of $\alpha$ to 0, 0.02, and 0.1, respectively.
\end{itemize}

\begin{table*}[t]
    \small
    \centering
    \vspace{-0.0cm}
    \caption{\textbf{Comparison of routing methods on RouterBench across three distinct performance-cost trade-off scenarios}. Bold and underline denote the best and second-best results. All methods are evaluated on Performance, Cost, and Score. The results are taken as the average of each dataset.}
    \setlength\tabcolsep{4.5pt}
    \begin{tabular}{p{2.5cm}<{\centering}p{1.1cm}<{\centering}p{1.1cm}<{\centering}p{1.1cm}<{\centering}p{1.1cm}<{\centering}p{1.1cm}<{\centering}p{1.1cm}<{\centering}p{1.1cm}<{\centering}p{1.1cm}<{\centering}p{1.1cm}<{\centering}}
        \toprule
        & \multicolumn{3}{c}{\textit{Performance First}} 
        & \multicolumn{3}{c}{\textit{Balance}} 
        & \multicolumn{3}{c}{\textit{Cost First}} \\
        \cmidrule(lr){2-4}\cmidrule(lr){5-7}\cmidrule(lr){8-10}

        & Perf.$\uparrow$ & Cost$\downarrow$ & Score$\uparrow$ & Perf.$\uparrow$ & Cost$\downarrow$ & Score$\uparrow$ & Perf.$\uparrow$ & Cost$\downarrow$ & Score$\uparrow$  \\
        
        \midrule
        \textit{Best candidate} & 0.813 & 7.185 & 0.813 & 0.709 & 0.562 & 0.698 & 0.704 & 0.439 & 0.660 \\ 
        
        \midrule
        Random           & 0.627 & 1.847 & 0.627 & 0.627 & 1.847 & 0.590 & 0.627 & 1.847 & 0.442 \\ 
        
        \midrule
        CosineClassifier & 0.662 & 1.448 & 0.662 & 0.584 & 0.189 & 0.580 & 0.566 & 0.162 & 0.549 \\ 
        HybridLLM        & 0.801 & 6.869 & 0.801 & 0.791 & 6.612 & 0.659 & 0.517 & 0.107 & 0.506 \\ 
        FrugalGPT        & 0.813 & 7.185 & 0.813 & 0.671 & 0.336 & 0.664 & 0.549 & 0.124 & 0.536 \\ 
        RouterDC         & 0.815 & 6.768 & \underline{0.815} & 0.716 & 1.313 & 0.690 & 0.718 & 0.418 & \underline{0.676} \\
        GraphRouter      & 0.813 & 7.185 & 0.813 & 0.713 & 0.987 & \underline{0.693} & 0.709 & 0.500 & 0.659 \\ \rowcolor{gray!20}
        RadialRouter     & 0.816 & 6.759 & \textbf{0.816} & 0.781 & 1.179 & \textbf{0.757} & 0.763 & 0.476 & \textbf{0.715} \\ 
        
        \midrule    
        Oracle           & 0.925 & 1.015 & 0.925 & 0.917 & 0.393 & 0.909 & 0.891 & 0.258 & 0.865 \\ 
        
        \bottomrule
    \end{tabular}
    \label{tab:comparison_results}
    \vspace{-0.4cm}
\end{table*}

\subsection{Implementation Details}

We adopt mDeBERTaV3-base \cite{he2021debertav3} as the language encoder $\mathcal{E}(\mathrm{x})$. For RadialFormer, the number of layers $T$ is set to 6, with a 768-dim hidden dimension. The head number of the multi-head attention is 4, with each head having a dimension of 32. The MLP for predicting the routing scores has a hidden layer dimension of 128. The training batch size is 64, and the maximum training epoch is 1000. The router is trained using the AdamW \cite{loshchilov2019decoupled} optimizer with a learning rate of $5\times10^{-5}$. The hyperparameter $\lambda$ is set to 0.5. All experiments are run on a single NVIDIA A100 80GB GPU.

\begin{table}[t]
    \small
    \centering
    \caption{\textbf{Ablation results on RadialRouter.} `\textit{PF}', `\textit{BA}', `\textit{CF}' denote three trade-off scenarios. $\mathcal{RF}$, Star-$\mathcal{T}$, $\mathcal{T}$ denote RadialFormer, Star-Transformer and standard Transformer, respectively. `Time' refers to the average routing time per batch in milliseconds. The best results are highlighted in \textbf{bold}.}
    \setlength\tabcolsep{4.5pt}
    \begin{tabular}{p{2.2cm}p{0.9cm}<{\centering}p{0.9cm}<{\centering}p{0.9cm}<{\centering}p{1.2cm}<{\centering}}
        \toprule
        Setting & \textit{PF} & \textit{BA} & \textit{CF} & Time/ms \\ \rowcolor{gray!20}
        \midrule
        RadialRouter & \textbf{0.816} & \textbf{0.757} & \textbf{0.715} & 10.7 \\ 
        \midrule
        \ $w/o$ $\mathcal{RF}$ \\ 
        \quad\ + Star-$\mathcal{T}$ & 0.813 & 0.751 & 0.709 & 13.5 \\ 
        \quad\ + $\mathcal{T}$      & 0.815 & 0.753 & 0.705 & 15.8 \\ 
        \quad\ + MLP                & 0.781 & 0.732 & 0.701 & \textbf{4.6} \\ 
        \midrule
        \ $w/o$ $\mathcal{L}_{\text{KL}}$  & 0.548 & 0.442 & 0.017 & - \\
        \ $w/o$ $\mathcal{L}_{\text{q-q}}$ & 0.813 & 0.740 & 0.711 & - \\ 
        \bottomrule
    \end{tabular}
    \label{tab:ablation_study}
    \vspace{-0.6cm}
\end{table}

\section{Empirical Results}

\subsection{Comparison with Baselines}

We compare RadialRouter with baselines in three scenarios. The results are shown in Tab.\ref{tab:comparison_results}. Here, `\textit{Best candidate}' denotes the individual LLM that achieves the highest score in the corresponding scenario. `Random' denotes randomly selecting LLMs from the LLM pool to generate responses to testing queries. We conduct 50 independent selections and calculate the average of the results obtained. `Oracle' denotes an ideal situation where all queries are routed to the optimal model, which defines the theoretical upper bound of the routing performance.

We can observe that RadialRouter substantially outperforms baseline methods in all three scenarios. In the \textit{Performance First} scenario, a relatively singular optimal LLM (GPT-4) yields similar performance outcomes across the routing methods. The routing process becomes complex considering the performance-cost trade-off, leading to greater disparities among the methods. RadialRouter surpasses baselines by at least 9.2\% and 5.8\% in the \textit{Balance} and \textit{Cost First} scenarios, demonstrating the superiority of the framework. The adaptability of RadialRouter to different performance-cost trade-offs is further verified in the Sec.\ref{sec:performance_cost}. RadialRouter significantly exceeds the \textit{Best candidate} and achieves at least 82.66\% of the Oracle's score. This suggests that RadialRouter is capable of implementing flexible routing within the LLM pool to improve the routing ability. In contrast, the baseline methods struggle with ineffective representation of the routing process, which limits their overall scores. This underscores the importance of discerning the intrinsic connection between the query and LLMs in the routing task.

\begin{figure}[t]
    \vspace{-0.4cm}
    \centering
    \subfloat[$w/o$\ $\mathcal{L_{\text{query-query}}}$.]{
    \includegraphics[width=0.225\textwidth]{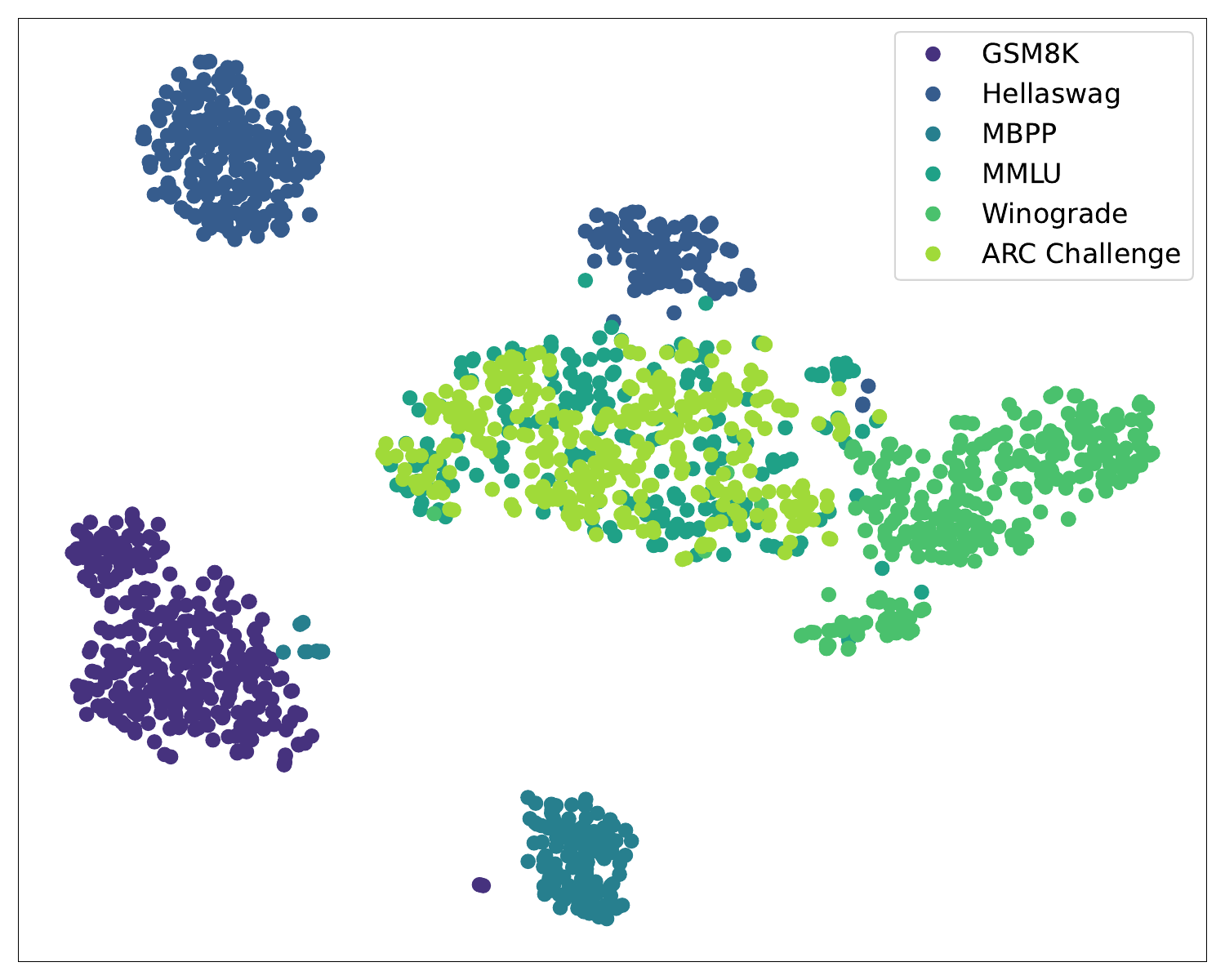}
    \label{fig:without_query_query_loss}}
    \hfill 
    \subfloat[$w/$\ $\mathcal{L_{\text{query-query}}}$.]{
    \includegraphics[width=0.225\textwidth]{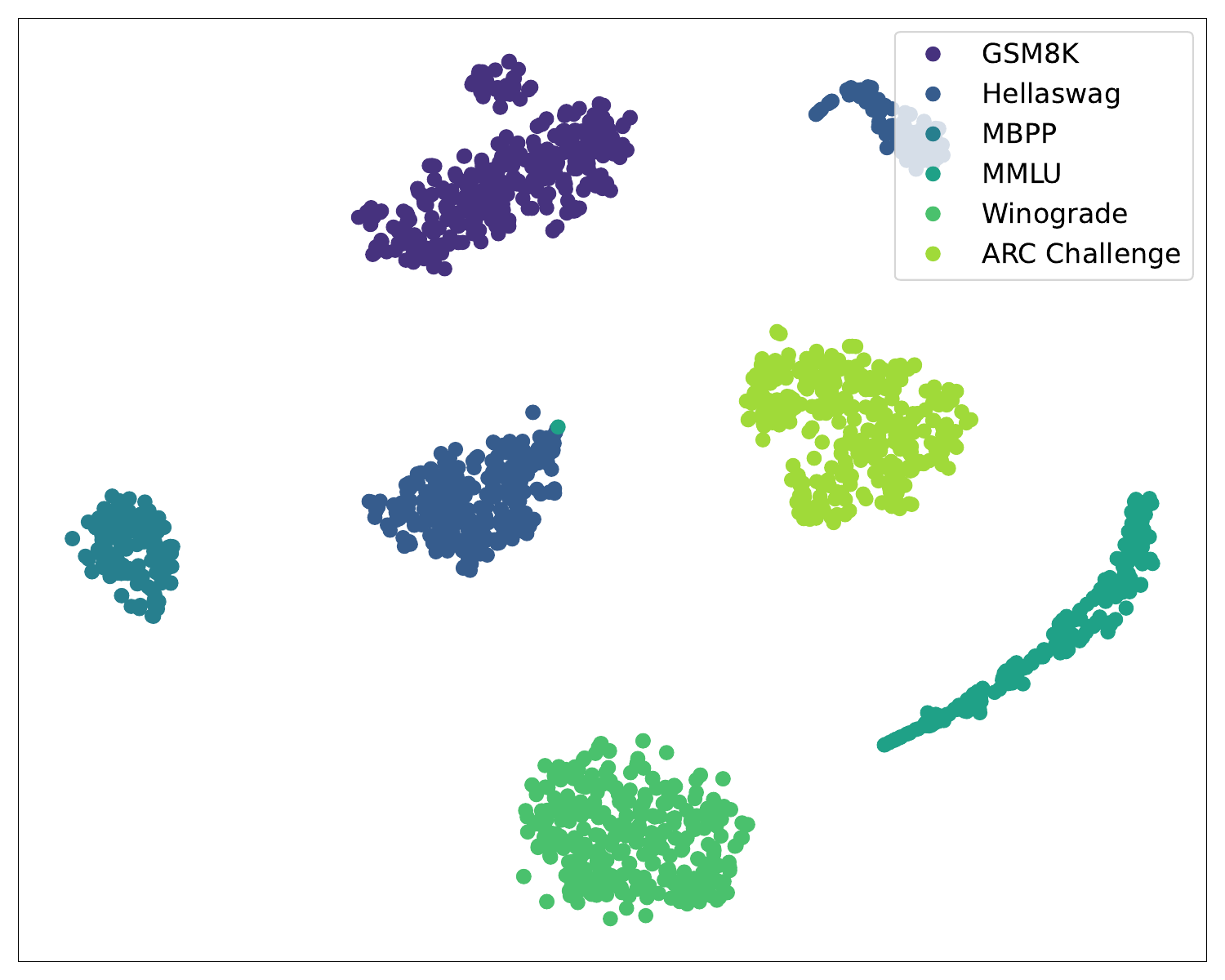}
    \label{fig:with_query_query_loss}}
    \caption{t-SNE visualization of test query embeddings extracted by the learned language encoder of RadialRouter.}
    \label{fig:tsne}
    \vspace{-0.6cm}
\end{figure}

\begin{figure*}[t]
    \centering
    \subfloat[Effects of $\alpha$ on routing methods. All methods are evaluated on Score.]{
    \includegraphics[width=0.42\textwidth]{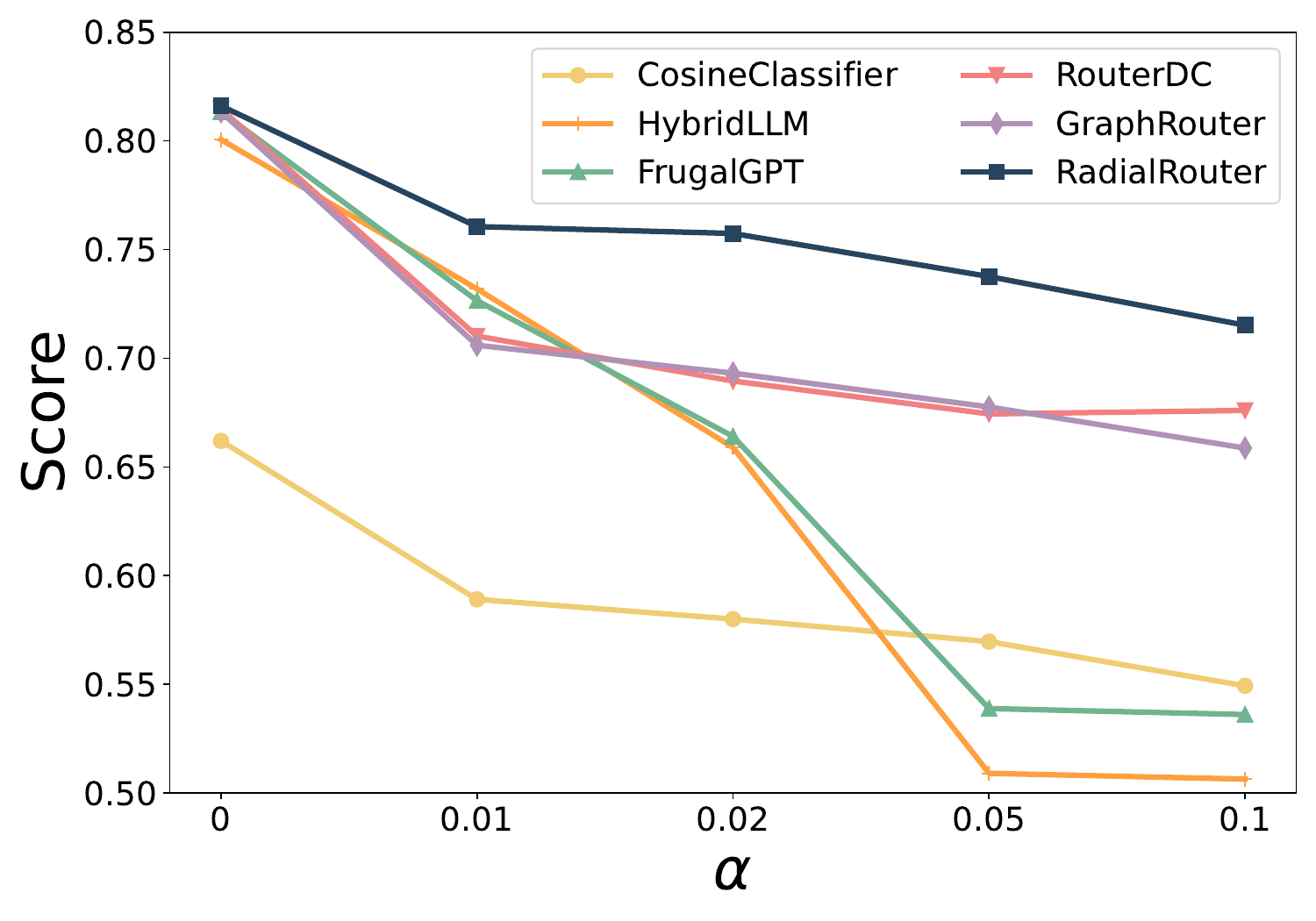}
    \label{fig:alpha_score}}
    \hspace{0.5cm} 
    \subfloat[Performance-cost trade-offs for routing methods. Scatter points denote candidate LLMs.]{
    \includegraphics[width=0.42\textwidth]{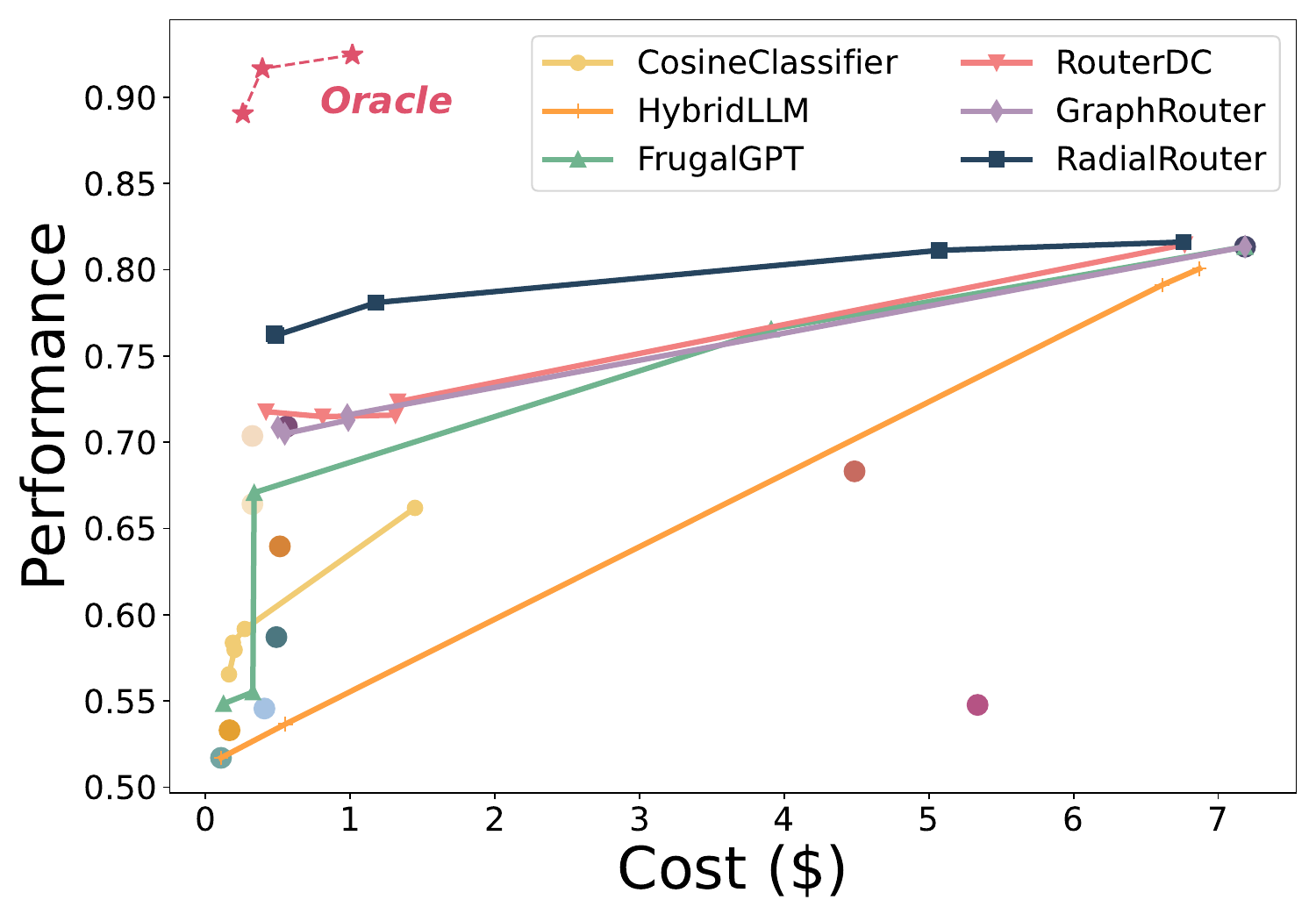}
    \label{fig:performance_cost}}
    \caption{Routing results on RouterBench within different performance-cost trade-offs.}
    \label{fig:performance_cost_trade_offs}
    \vspace{-0.4cm}
\end{figure*}

\subsection{Ablation Studies}

We conduct ablation studies to investigate the contribution of each component in RadialRouter as shown in Tab.\ref{tab:ablation_study}. Here, `$w/o$' denotes variants with specific components removed, and `$+$' denotes replacing RadialFormer with alternative architectures.

We model the routing problem as a probability distribution prediciton problem which requires efficient representation of both queries and LLMs. From Tab.\ref{tab:ablation_study} we can observe that replacing RadialFormer with alternative architectures (Star-Transformer \cite{guo2019star}, standard Transformer \cite{vaswani2017attention}, and MLP) all leads to performance degradation. The result indicates the effectiveness of RadialFormer, which maintains rich semantic information through structured representation and facilitates the selection of the optimal LLM. In construct, the architecture of Star-Transformer introduces unnecessary connections (e.g., Ring Connections), causing interference to the routing process. We further compare the routing efficiencies of different architectures. RadialFormer demonstrates lower time consumption than both Star-Transformer and standard Transformer, validating its lightweight design.

Eliminating the KL divergence loss leads to a substantial drop in metrics, further experiments in Sec.\ref{sec:loss_function_for_llm_selection} investigate the impact of loss functions on the optimal LLM selection. Eliminating the query-query contrastive loss also leads to a performance decline. Fig.\ref{fig:tsne} exhibits the t-SNE visualization of test query embeddings extracted by the learned language encoder of RadialRouter. We can observe that the absence of the query-query contrastive loss results in mixed query embeddings across different datasets. By incorporating the contrastive loss, we achieve well-separated query embeddings, thereby establishing a robust foundation for effective routing. Detailed results of ablation studies are shown in Tab.\ref{tab:detailed_ablation_study} of Appendix \ref{sec:detailed_results_of_ablation_study}.

\begin{table}[t]
    \small
    \centering
    \caption{\textbf{Comparison on different loss functions for LLM selection.} `\textit{PF}', `\textit{BA}', `\textit{CF}' denote three trade-off scenarios. All settings are evaluated on Score. The best results are highlighted in \textbf{bold}.}
    \setlength\tabcolsep{4.5pt}
    \begin{tabular}{p{3cm}p{1cm}<{\centering}p{1cm}<{\centering}p{1cm}<{\centering}}
        \toprule
        Model Setting & \textit{PF} & \textit{BA} & \textit{CF} \\ \rowcolor{gray!20}
        \midrule
        RadialRouter $w/\ \mathcal{L}_{\text{KL}}$  & \textbf{0.816} & \textbf{0.757} & \textbf{0.715}  \\ 
        RadialRouter $w/\ \mathcal{L}_{\text{ce}}$  & 0.533 & 0.530 & 0.520  \\  
        RadialRouter $w/\ \mathcal{L}_{\text{q-L}}$ & 0.815 & 0.714 & 0.676  \\ 
        \bottomrule
    \end{tabular}
    \label{tab:loss_functions_for_llm_selection}
    \vspace{-0.2cm}
\end{table}

\begin{table*}[t]
    \small
    \centering
    \caption{\textbf{Comparison on routing to an increasing number of candidate LLMs} in the \textit{Balance} scenario, where `+' denotes an increase in the LLM pool. The results are taken as the average of each dataset.}
    \setlength\tabcolsep{4.5pt}
    \begin{tabular}{p{4cm}p{1.5cm}<{\centering}p{2.2cm}<{\centering}p{2cm}<{\centering}p{2cm}<{\centering}}
        \toprule
        & $\#$LLM & \textbf{Performance}$\uparrow$ & \textbf{Cost}$\downarrow$ & \textbf{Score}$\uparrow$ \\  
        \midrule
        WizardLM-13B-V1.2                        & 1  & 0.5331 & 0.166 & 0.530 \\ 
        \ + code-llama-34b-chat                  & 2  & 0.5539 & 0.178 & 0.550 \\ 
        \ \ + llama-2-70b-chat                   & 3  & 0.6105 & 0.468 & 0.601 \\ 
        \ \ \ + claude-v2                        & 4  & 0.6550 & 2.348 & 0.608 \\ 
        \ \ \ \ + claude-v1                      & 5  & 0.6696 & 2.758 & 0.614 \\ 
        \ \ \ \ \ + claude-instant-v1            & 6  & 0.6731 & 1.134 & 0.650 \\ 
        \ \ \ \ \ \ + mistral-7b-chat            & 7  & 0.6731 & 1.134 & 0.650 \\ 
        \ \ \ \ \ \ \ + mixtral-8x7b-chat        & 8  & 0.6769 & 1.109 & 0.655 \\ 
        \ \ \ \ \ \ \ \ + Yi-34B-Chat            & 9  & 0.6964 & 0.421 & 0.688 \\ 
        \ \ \ \ \ \ \ \ \ + gpt-3.5-turbo-1106   & 10 & 0.7068 & 0.404 & 0.699 \\ \rowcolor{gray!20}
        \ \ \ \ \ \ \ \ \ \ + gpt-4-1106-preview & 11 & 0.7810 & 1.179 & 0.757 \\  
        \bottomrule
    \end{tabular}
    \label{tab:different_numbers}
    \vspace{-0.2cm}
\end{table*}

\begin{figure}[t]
  \centering
  \setlength{\abovecaptionskip}{0.2cm}
  \includegraphics[width=0.42\textwidth]{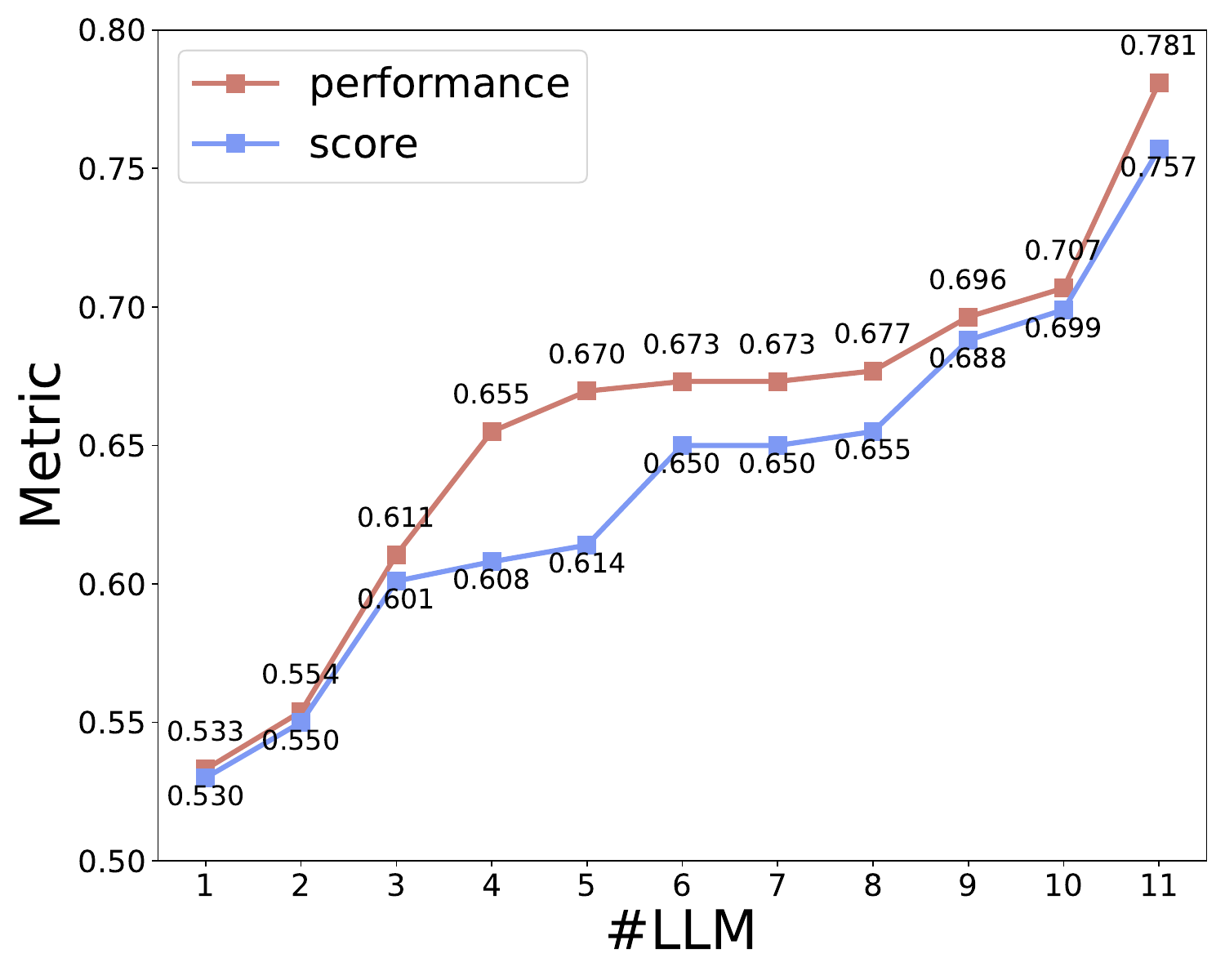}
  \caption{Effects of different numbers of LLMs.}
  \label{fig:number_of_llms}
  \vspace{-0.5cm}
\end{figure}

\subsection{Analysis}

\subsubsection{Loss Function for LLM Selection}
\label{sec:loss_function_for_llm_selection}

We further compare the Kullback-Leibler divergence loss with two different loss functions for supervision the optimal LLM selection.

\paragraph{Cross-Entropy Loss} 

Viewing LLM routing as a multi-class classification problem, the cross-entropy loss is introduced. In this approach, the LLM that receives the highest true score is assigned `1', while all other LLMs are assigned `0'. The cross-entropy loss function is defined as:
\begin{equation}
    \mathcal{L}_{\text{ce}}(\mathrm{x};\bm{\theta})=-\sum_{i=1}^{n}{y_i\log(p_i)},
\end{equation}
where $y_i$ denotes the label for $\mathrm{LLM}_i$, and $p$ denotes the predicted routing probability.

\paragraph{Query-LLM Contrastive Loss}

Considering the objective of routing is to allocate the query to top-performing LLMs, rather than merely identifying the optimal model, \cite{chen2024routerdc} introduces the sample-LLM contrastive loss to learn the router. We make minor modifications to it within the RadialRouter framework. Based on the true scores obtained by Eq.\ref{eq:score}, we construct the LLMs index set $\mathcal{I}$ and $\bar{\mathcal{I}}$ as the indices of LLMs corresponding to the top-$K$ and bottom-$K$ scores, respectively. The query-LLM contrastive loss is then defined as:
\begin{equation}
    \mathcal{L}_{\text{q-L}}(\mathrm{x};\bm{\theta})=\sum_{i\in\mathcal{I}}-\log\frac{e^{p_{i}}}{e^{p_{i}}+\sum\limits_{j\in\bar{\mathcal{I}}}e^{p_j}},
\end{equation}
where $p$ denotes the predicted routing probability, $i$ and $j$ denote the index of LLMs corresponding to the top-$K$ and bottom-$K$ scores, respectively.

Tab.\ref{tab:loss_functions_for_llm_selection} displays the results of comparison. Training RadialRouter with  $\mathcal{L}_{\text{KL}}$ yields the highest scores in all three scenarios, as can be observed. From a macroscopic perspective, $\mathcal{L}_{\text{KL}}$, $\mathcal{L}_{\text{ce}}$ and $\mathcal{L}_{\text{q-L}}$ address the routing problem out of distinct viewpoints: probabilistic distribution fitting, multi-class classification, and contrastive learning, respectively. The superiority of the KL divergence underscores the effectiveness of framing the routing problem through the lens of probabilistic distribution fitting, which fosters a comprehensive understanding of the intrinsic connection between the query and LLMs in the routing process, as opposed to focusing merely on a limited number of dominant LLMs. This perspective aligns seamlessly with the design principles of RadialFormer, allowing for the accomplishment of both efficient and robust routing.

\subsubsection{Adaptability to Performance-Cost Trade-Offs}
\label{sec:performance_cost}

Beyond the three aforementioned scenarios, we assess the adaptability of RadialRouter and baseline routing methods to performance-cost trade-offs by varying the parameter $\alpha$ in Eq.\ref{eq:score}. Fig.\ref{fig:alpha_score} shows the scores achieved by routing methods within different $\alpha$, where RadialRouter achieves the highest scores in different trade-offs, significantly outperforming baseline routing methods. Fig.\ref{fig:performance_cost} depicts the performance-cost curves of routing methods. Our observation indicate that RadialRouter is capable to obtain improved performance while maintaining comparable costs to baselines, leading to a robust performance-cost balance. This analysis substantiates the adaptability of RadialRouter to performance-cost trade-offs, which highlights its practical applicability. Detailed results are shown in Tab.\ref{tab:full_results} of Appendix \ref{sec:full_results_of_performance_cost_balance}.

\subsubsection{Routing to Different Numbers of LLMs}

We evaluate the efficiency of RadialRouter with a dynamic LLM pool by gradually increasing the number of candidate LLMs. We compare the results in the \textit{Balance} scenario, as shown in Tab.\ref{tab:different_numbers} and Fig.\ref{fig:number_of_llms}. We can observe that increasing the number of candidate LLMs leads to improved performance, demonstrating RadialRouter's ability to effectively adapt to a dynamic LLM pool.

\subsubsection{Effects of $\lambda$ on Optimization}

We study the impact of the contrastive loss weight $\lambda$ in Eq.\ref{eq:optimization} on the optimization of RadialRouter. Experiments are conducted in the \textit{Balance} scenario, and the results are presented in Appendix \ref{sec:full_results_of_effects_of_lambda}. As can be seen, the highest score is achieved when $\lambda=0.5$. Therefore, we fix $\lambda$ to 0.5 when training RadialRouter. Moreover, we observe that RadialRouter demonstrates insensitivity across a broad range of $\lambda\in[0.25,5]$, which confirms the robustness of our method and provides greater flexibility in the practical selection of $\lambda$.

\section{Conclusions}

In this paper, we introduce RadialRouter, a novel Transformer-based framework for efficient and robust LLM routing. We achieve a structured representation of the routing process with RadialFormer, which efficiently captures the interrelationship between the query and LLMs. The robustness of RadialRouter is enhanced by incorporating contrastive loss. Extensive experiments on RouterBench demonstrate that RadialRouter consistently outperforms baseline methods across various scenarios, exhibiting adaptability to performance-cost trade-offs and efficacy in routing queries within a dynamic LLM pool. These findings confirm the potential of RadialRouter as a superior solution for LLM deployment in practical applications.

\section*{Limitations}

We acknowledge several limitations regarding our proposed method. 

First, RadialRouter requires re-training whenever a new LLM is introduced to the LLM pool. Consequently, the ability to rapidly adapt and iterate in dynamic environments is hindered, particularly in scenarios where frequent updates are needed to address evolving tasks or domains. The router's implementation of training-free adaptation to the dynamic LLM pool relies on a general portrait or embedding for various LLMs, but that is beyond the scope of this paper.

Second, due to limited computational resources, we have not performed testing within multi-language and multi-modal LLM ensembles, which may restrict our ability to fully assess the framework's applicability across languages and modalities. We leave the investigation of such scenarios to future work.

\bibliography{custom}
\bibliographystyle{acl_natbib}

\newpage
\appendix

\section{Details of Multi-Head Attention}
\label{sec:multi_head_attention}

The update of RadialFormer is based on the attention mechanism \cite{vaswani2017attention}. Specifically, given the input sequence $\mathbf{H}\in\mathbb{R}^{m\times d}$ where $m$ denotes the sequence length and $d$ denotes the hidden dimension, we use a query sequence $\mathbf{q}\in\mathbb{R}^{1\times d}$ to compute the relevant information based on the scaled dot-product attention:
\begin{equation}
    \mathrm{Attn}(Q,K,V)=\mathrm{softmax}(\frac{QK^\mathsf{T}}{\sqrt{d}})V,
\end{equation}
where $[Q;K;V]=[\mathbf{q}W_q;\mathbf{H}W_k;\mathbf{H}W_v]$, and $W_q,W_k,W_v$ denote learnable parameters. 

The multi-head attention layer extends the scale dot-product attention by performing $h$ paralleled attention operations and concatenating the information:
\begin{equation}
\mathrm{MHAttn}(\mathbf{q},\mathbf{H})=[\mathrm{head}_1,\ldots,\mathrm{head}_h]W_o,
\end{equation}
\begin{equation}
\mathrm{head}_i=\mathrm{Attn}(Q_i,K_i,V_i),i\in[1,h],
\end{equation}
where $[Q_i;K_i,;V_i]$ are the $i$-th group from $[Q;K;V]$ with a dimension of $d/h$, and $W_o$ denotes output learnable parameter.

In RadialFormer, the multi-head attention mechanism is utilized to update states of the relay node and the satellite nodes.

\section{Training and Inference Algorithm of RadialRouter}
\label{sec:training_and_inference_algorithm_of_radialrouter}

Alg.\ref{alg:radialrouter} shows the training and inference procedures of RadialRouter. The algorithm presented in RadialRouter outlines a training and inference framework for efficient and robust LLM routing. During training, the learnable parameters in the router are updated through mini-batch sampling, supervised by the Kullback-Leibler divergence loss and the query-query contrastive loss. Given an input query, the inference of RadialRouter involves predicting the routing probabilities for the candidate LLMs and selecting the optimal one for response.

\begin{algorithm*}[t]
    \caption{The overall algorithm of \textbf{RadialRouter}}
    \label{alg:radialrouter}
    \begin{algorithmic}[1]
    \renewcommand{\algorithmicrequire}{\textbf{Input:}} 
    \renewcommand{\algorithmicensure}{\textbf{Output:}}
    \Require training set $\mathcal{D}_{\text{train}}$, LLMs $\{\mathrm{LLM}_i:i=1,\dots,n\}$, number of out-group queries $H$, number of clusters $N$, hyper-parameter $\lambda$, mini-batch size $b$, and learning rate $\eta$; learnable parameters $\bm{\theta}$: encoder $\mathcal{E}$, RadialFormer $\mathcal{RF}$, MLP $\mathcal{M}$, and learnable LLM embeddings $\{\mathbf{m}_i:i=1,\dots,n\}$;
    \Statex
    \Statex \textit{Training:}
    \State Score LLMs for each query $(\mathrm{x}_j,\mathrm{y}_j)\in\mathcal{D}_{\text{train}}$ and obtain $\{s_j^{(i)}:i=1,\dots,n\}$ by Eq.\ref{eq:score};
    \State Cluster training queries $\{\mathrm{x}_j:j=1,\dots,l\}$ into $N$ groups $\{\mathcal{K}_1,\ldots,\mathcal{K}_N\}$;
    \Repeat
    \State Sample a mini-batch $\mathcal{B}$ from $\mathcal{D}_{\text{train}}$;
    \For{$(\mathrm{x}_j,\mathrm{y}_j)\in\mathcal{B}$}
    \State $\mathbf{q}\gets\mathcal{E}(\mathrm{x}_j)$;
    \State Compute the updated state of RadialFormer by Alg.\ref{alg:radialformer}: 
    \State $\br^T,\bs_1^T,\ldots,\bs_n^T\gets\mathcal{RF}(\mathbf{q},\mathbf{m}_1,\ldots,\mathbf{m}_n)$;
    \State Compute the Kullback-Leibler divergence loss $\mathcal{L}_{\text{Kullback-Leibler}}(\mathrm{x}_j;\bm{\theta})$ by Eq.\ref{eq:kullback-leibler};
    \State Sample an in-group query $\mathrm{x}^+$ and $H$ out-group queries $\mathrm{x}_t^-$ from $\mathcal{B}$;
    \State Compute the query-query contrastive loss $\mathcal{L}_{\text{query-query}}(\mathrm{x}_i;\bm{\theta})$ by Eq.\ref{eq:query_query};
    \EndFor
    \State $\mathcal{L}(\mathcal{B};\bm{\theta})\gets\sum_{\mathrm{x}_i\in\mathcal{B}}\mathcal{L}_{\text{Kullback-Leibler}}(\mathrm{x}_i;\bm{\theta})+\lambda\mathcal{L}_{\text{query-query}}(\mathrm{x}_i;\bm{\theta})$;
    \State $\bm{\theta}\gets\bm{\theta}-\eta\nabla_{\bm{\theta}}\mathcal{L}(\mathcal{B};\bm{\theta})$;
    \Until{converged.}
    \Statex
    \Statex \textit{Inference:}
    \State Sample a testing query $\mathrm{x}'$;
    \State Compute the predicted routing probability $p'$ using the $\mathcal{E}$, $\mathcal{RF}$ and $\mathcal{M}$;
    \State $i'\gets\arg\max_{i\in\{1,\dots,n\}}(p_i')$;
    \State $\hat{\mathrm{y}}'\gets\mathrm{LLM}_{i'}(\mathrm{x}')$.
    \Ensure response $\hat{\mathrm{y}}'$
    \end{algorithmic}
\end{algorithm*}

\section{Statistics of Candidate LLMs on RouterBench}
\label{sec:statistics_of_candidate_llms}

Tab.\ref{tab:statistics_llms} shows the basic statistics of 11 LLMs in RouterBench as our candidate LLMs. Tab.\ref{tab:candidate_llms} shows the statistics of candidate LLMs on RouterBench considering performance and cost. The performance of each candidate LLM is assessed by averaging the accuracy across six different datasets. The cost is determined based on the pricing of the LLMs per million tokens, and is also averaged over the six datasets.

\begin{table}[t]
    \small
    \centering
    \vspace{-0.4cm}
    \caption{Statistics of different LLMs in RouterBench.}
    \setlength\tabcolsep{4.5pt}
    \begin{tabular}{p{0.6cm}<{\centering}p{3.2cm}p{1cm}<{\centering}}
        \toprule
        & \textbf{LLM} & \textbf{Size} \\
        \midrule
        \multirow{6}*{\rotatebox{90}{\textit{open-source}}} & WizardLM-13B-V1.2   & 13B \\
                                                            & code-llama-34b-chat & 34B \\
                                                            & llama-2-70b-chat    & 70B \\
                                                            & mistral-7b-chat     & 7B  \\
                                                            & mixtral-8x7b-chat   & 47B \\
                                                            & Yi-34B-Chat         & 34B \\
        \midrule
        \multirow{5}*{\rotatebox{90}{\textit{proprietary}}} & claude-instant-v1   & - \\
                                                            & claude-v1           & - \\
                                                            & claude-v2           & - \\
                                                            & gpt-3.5-turbo-1106  & - \\
                                                            & gpt-4-1106-preview  & - \\
        \bottomrule
    \end{tabular}
    \label{tab:statistics_llms}
    \vspace{-0.5cm}
\end{table}

\begin{table}[t]
    \small
    \centering
    \vspace{-0.4cm}
    \caption{\textbf{Effects of $\lambda$} in the \textit{Balance} scenario. The best results are highlighted in \textbf{bold}.}
    \setlength\tabcolsep{4.5pt}
    \begin{tabular}{p{0.8cm}<{\centering}p{1.9cm}<{\centering}p{1.3cm}<{\centering}p{1.3cm}<{\centering}}
        \toprule
        $\lambda$ & \textbf{Performance} & \textbf{Cost} & \textbf{Score} \\
        \midrule
        0    & 0.7492 & 0.455 & 0.740  \\ 
        0.25 & 0.7911 & 1.964 & 0.752  \\ \rowcolor{gray!20}
        0.5  & 0.7810 & 1.179 & \textbf{0.757}  \\ 
        0.75 & \textbf{0.7923} & 1.960 & 0.753  \\ 
        1    & 0.7911 & 1.964 & 0.752  \\  
        2    & 0.7845 & 1.985 & 0.745  \\ 
        3    & 0.7911 & 1.964 & 0.752  \\ 
        4    & 0.7589 & 0.478 & 0.749  \\ 
        5    & 0.7620 & 0.488 & 0.752  \\ 
        6    & 0.7577 & 0.482 & 0.748  \\ 
        8    & 0.7581 & 0.481 & 0.748  \\ 
        10   & 0.7575 & 0.481 & 0.748  \\ 
        \bottomrule
    \end{tabular}
    \label{tab:effects_of_lambda}
    \vspace{-0.4cm}
\end{table}

\begin{table*}[t]
    \centering
    \caption{Statistics of candidate LLMs on RouterBench.}
    \setlength\tabcolsep{4.2pt}
    \begin{tabular}{ccccccccc}
        \toprule
        \textbf{LLM} & GSM8K & Hellaswag & MBPP & MMLU & winograde & ARC & \textbf{Perf.}$\uparrow$ & \textbf{Cost}$\downarrow$ \\ 
        
        \midrule
        WizardLM-13B-V1.2 & 0.5054 & 0.6004 & 0.3906 & 0.5253 & 0.5289 & 0.6476 & 0.5331 & 0.166 \\ 
        claude-instant-v1 & 0.6281 & 0.7690 & 0.6250 & 0.4529 & 0.5211 & 0.8421 & 0.6397 & 0.514 \\ 
        claude-v1 & 0.6520 & 0.8187 & 0.6094 & 0.5281 & 0.5711 & 0.9199 & 0.6832 & 4.486 \\ 
        claude-v2 & 0.6671 & 0.3130 & 0.6406 & 0.5652 & 0.4763 & 0.6247 & 0.5478 & 5.336 \\ 
        gpt-3.5-turbo-1106 & 0.6094 & 0.7843 & 0.6875 & 0.6667 & 0.6632 & 0.8444 & 0.7092 & 0.562 \\ 
        gpt-4-1106-preview & 0.6589 & 0.9057 & 0.6875 & 0.8162 & 0.8552 & 0.9565 & 0.8134 & 7.185 \\ 
        code-llama-34b-chat & 0.4548 & 0.5194 & 0.5156 & 0.5284 & 0.5921 & 0.6636 & 0.5457 & 0.407 \\ 
        llama-2-70b-chat & 0.5252 & 0.7046 & 0.3750 & 0.6034 & 0.4974 & 0.8169 & 0.5871 & 0.490 \\ 
        mistral-7b-chat & 0.4151 & 0.5410 & 0.3828 & 0.5198 & 0.5737 & 0.6705 & 0.5171 & 0.107 \\ 
        mixtral-8x7b-chat & 0.5214 & 0.6960 & 0.5391 & 0.6822 & 0.6842 & 0.8627 & 0.6642 & 0.324 \\ 
        Yi-34B-Chat & 0.5517 & 0.8782 & 0.4141 & 0.7187 & 0.7421 & 0.9176 & 0.7037 & 0.439 \\ 
        
        \bottomrule
    \end{tabular}
    \label{tab:candidate_llms}
    \vspace{-0cm}
\end{table*}

\section{Detailed Results of Ablation Studies}
\label{sec:detailed_results_of_ablation_study}

Tab.\ref{tab:detailed_ablation_study} shows the detailed results of ablation studies on RadialRouter. Through ablation studies, we verify the rationality of RadialFormer and the necessity of introducing the KL divergence loss and the query-query contrastive loss.

\begin{table*}[t]
    \centering
    \caption{\textbf{Detailed ablation results on RadialRouter.} The best results are highlighted in \textbf{bold}.}
    \setlength\tabcolsep{4.5pt}
    \begin{tabular}{p{3cm}p{1cm}<{\centering}p{1cm}<{\centering}p{1cm}<{\centering}p{1cm}<{\centering}p{1cm}<{\centering}p{1cm}<{\centering}p{1cm}<{\centering}p{1cm}<{\centering}p{1cm}<{\centering}}
        \toprule
        & \multicolumn{3}{c}{\textit{Performance First}} 
        & \multicolumn{3}{c}{\textit{Balance}} 
        & \multicolumn{3}{c}{\textit{Cost First}} \\
        \cmidrule(lr){2-4}\cmidrule(lr){5-7}\cmidrule(lr){8-10}

        & Perf. & Cost & Score & Perf. & Cost & Score & Perf. & Cost & Score \\ \rowcolor{gray!20}
        
        \midrule
        RadialRouter & 0.816 & 6.759 & \textbf{0.816} & 0.781 & 1.179 & \textbf{0.757} & 0.763 & 0.476 & \textbf{0.715} \\ 
        \midrule
        $w/o$ RadialFormer \\ 
        \ + Star-Transformer & 0.813 & 7.185 & 0.813 & 0.794 & 2.170 & 0.751 & 0.758 & 0.491 & 0.709 \\ 
        \ + Transformer      & 0.815 & 6.768 & 0.815 & 0.792 & 1.960 & 0.753 & 0.752 & 0.478 & 0.705 \\ 
        \ + MLP              & 0.781 & 4.362 & 0.781 & 0.770 & 1.940 & 0.732 & 0.751 & 0.496 & 0.701 \\ 
        \midrule
        $w/o\ \mathcal{L}_{\text{KL}}$  & 0.548 & 5.308 & 0.548 & 0.548 & 5.308 & 0.442 & 0.548 & 5.308 & 0.017 \\
        $w/o\ \mathcal{L}_{\text{q-q}}$ & 0.813 & 7.185 & 0.813 & 0.759 & 0.519 & 0.749 & 0.759 & 0.478 & 0.711 \\ 
        \bottomrule
    \end{tabular}
    \label{tab:detailed_ablation_study}
    \vspace{-0.4cm}
\end{table*}

\section{Full Results of Performance-Cost Balance}
\label{sec:full_results_of_performance_cost_balance}

Tab.\ref{tab:full_results} is the full results of Fig.\ref{fig:performance_cost_trade_offs}. We set the value of $\alpha$ to 0, 0.01, 0.02, 0.05, and 0.1 to assess the adaptability of different routing methods to performance-cost trade-offs. We can see that RadialRouter is robust to a wide range of performance-cost trade-off scenarios ($\alpha\in[0,0.1]$).

\begin{table*}[t]
    \centering
    \caption{Performance and cost of routing methods on RouterBench with different $\alpha$.}
    \setlength\tabcolsep{4.2pt}
    \begin{tabular}{cccccccccc}
        \toprule
        \textbf{Method} & $\alpha$ & GSM8K & Hellaswag & MBPP & MMLU & winograde & ARC & \textbf{Perf.}$\uparrow$ & \textbf{Cost}$\downarrow$ \\
        
        \midrule
        \multirow{4}*{CosineClassifier} & 0 & 0.6062 & 0.7046 & 0.6250 & 0.6798 & 0.5395 & 0.8169 & 0.6620 & 1.448 \\ 
        & 0.01 & 0.4671 & 0.6004 & 0.4922 & 0.5730 & 0.5553 & 0.8627 & 0.5918 & 0.271 \\
        & 0.02 & 0.4725 & 0.5410 & 0.4766 & 0.5765 & 0.5737 & 0.8627 & 0.5838 & 0.189 \\
        & 0.05 & 0.4320 & 0.5410 & 0.5000 & 0.5663 & 0.5763 & 0.8627 & 0.5797 & 0.201 \\
        & 0.1  & 0.4945 & 0.5410 & 0.5703 & 0.5352 & 0.5816 & 0.6705 & 0.5655 & 0.162 \\
        \midrule
        \multirow{4}*{HybridLLM} & 0 & 0.6489 & 0.8898 & 0.6719 & 0.8054 & 0.8474 & 0.9405 & 0.8006 & 6.869 \\ 
        & 0.01 & 0.6489 & 0.8898 & 0.6719 & 0.8054 & 0.8474 & 0.9405 & 0.8006 & 6.869 \\
        & 0.02 & 0.6371 & 0.8715 & 0.6719 & 0.7904 & 0.8474 & 0.9291 & 0.7912 & 6.612 \\
        & 0.05 & 0.4294 & 0.5586 & 0.4219 & 0.5393 & 0.5868 & 0.6842 & 0.5367 & 0.553 \\
        & 0.1  & 0.4151 & 0.5410 & 0.3828 & 0.5198 & 0.5737 & 0.6705 & 0.5171 & 0.107 \\
        \midrule
        \multirow{4}*{FrugalGPT} & 0 & 0.6589 & 0.9057 & 0.6875 & 0.8162 & 0.8552 & 0.9565 & 0.8134 & 7.185 \\ 
        & 0.01 & 0.6317 & 0.8437 & 0.6953 & 0.7422 & 0.7816 & 0.8993 & 0.7656 & 3.910 \\
        & 0.02 & 0.5229 & 0.7172 & 0.5312 & 0.6888 & 0.6947 & 0.8696 & 0.6708 & 0.336 \\
        & 0.05 & 0.5056 & 0.6004 & 0.5234 & 0.5253 & 0.5289 & 0.6476 & 0.5552 & 0.327 \\
        & 0.1  & 0.4154 & 0.5410 & 0.5703 & 0.5198 & 0.5737 & 0.6705 & 0.5485 & 0.124 \\
        \midrule 
        \multirow{4}*{RouterDC} & 0 & 0.6671 & 0.9057 & 0.6875 & 0.8163 & 0.8553 & 0.9565 & 0.8147 & 6.768 \\ 
        & 0.01 & 0.6671 & 0.8782 & 0.5078 & 0.6860 & 0.6842 & 0.9176 & 0.7235 & 1.329 \\
        & 0.02 & 0.6671 & 0.8782 & 0.5156 & 0.6869 & 0.6842 & 0.8627 & 0.7158 & 1.313 \\
        & 0.05 & 0.6094 & 0.8782 & 0.6016 & 0.5822 & 0.7553 & 0.8627 & 0.7149 & 0.810 \\
        & 0.1  & 0.6281 & 0.8782 & 0.5078 & 0.6910 & 0.6842 & 0.9176 & 0.7178 & 0.418 \\
        \midrule
        \multirow{4}*{GraphRouter} & 0 & 0.6589 & 0.9057 & 0.6875 & 0.8162 & 0.8552 & 0.9565 & 0.8134 & 7.185 \\
        & 0.01 & 0.6121 & 0.7902 & 0.6953 & 0.6751 & 0.6711 & 0.8513 & 0.7158 & 0.980 \\
        & 0.02 & 0.6121 & 0.7886 & 0.6875 & 0.6739 & 0.6711 & 0.8444 & 0.7129 & 0.987 \\
        & 0.05 & 0.6013 & 0.7896 & 0.6719 & 0.6698 & 0.6553 & 0.8421 & 0.7050 & 0.548 \\
        & 0.1  & 0.5802 & 0.8321 & 0.5859 & 0.6926 & 0.6895 & 0.8719 & 0.7087 & 0.500 \\
        \midrule
        \multirow{4}*{RadialRouter} & 0 & 0.6672 & 0.9057 & 0.6953 & 0.8163 & 0.8553 & 0.9565 & 0.8161 & 6.759 \\
        & 0.01 & 0.6671 & 0.8782 & 0.7031 & 0.8077 & 0.8553 & 0.9565 & 0.8113 & 5.072 \\
        & 0.02 & 0.6281 & 0.8782 & 0.6797 & 0.7270 & 0.8553 & 0.9176 & 0.7810 & 1.179 \\
        & 0.05 & 0.6281 & 0.8782 & 0.6875 & 0.7187 & 0.7421 & 0.9176 & 0.7620 & 0.488 \\
        & 0.1  & 0.6281 & 0.8782 & 0.6875 & 0.7235 & 0.7421 & 0.9176 & 0.7628 & 0.476 \\
        
        \bottomrule
    \end{tabular}
    \label{tab:full_results}
    \vspace{-0.4cm}
\end{table*}

\begin{figure}[t]
  \centering
  \setlength{\abovecaptionskip}{0.2cm}
  \includegraphics[width=0.48\textwidth]{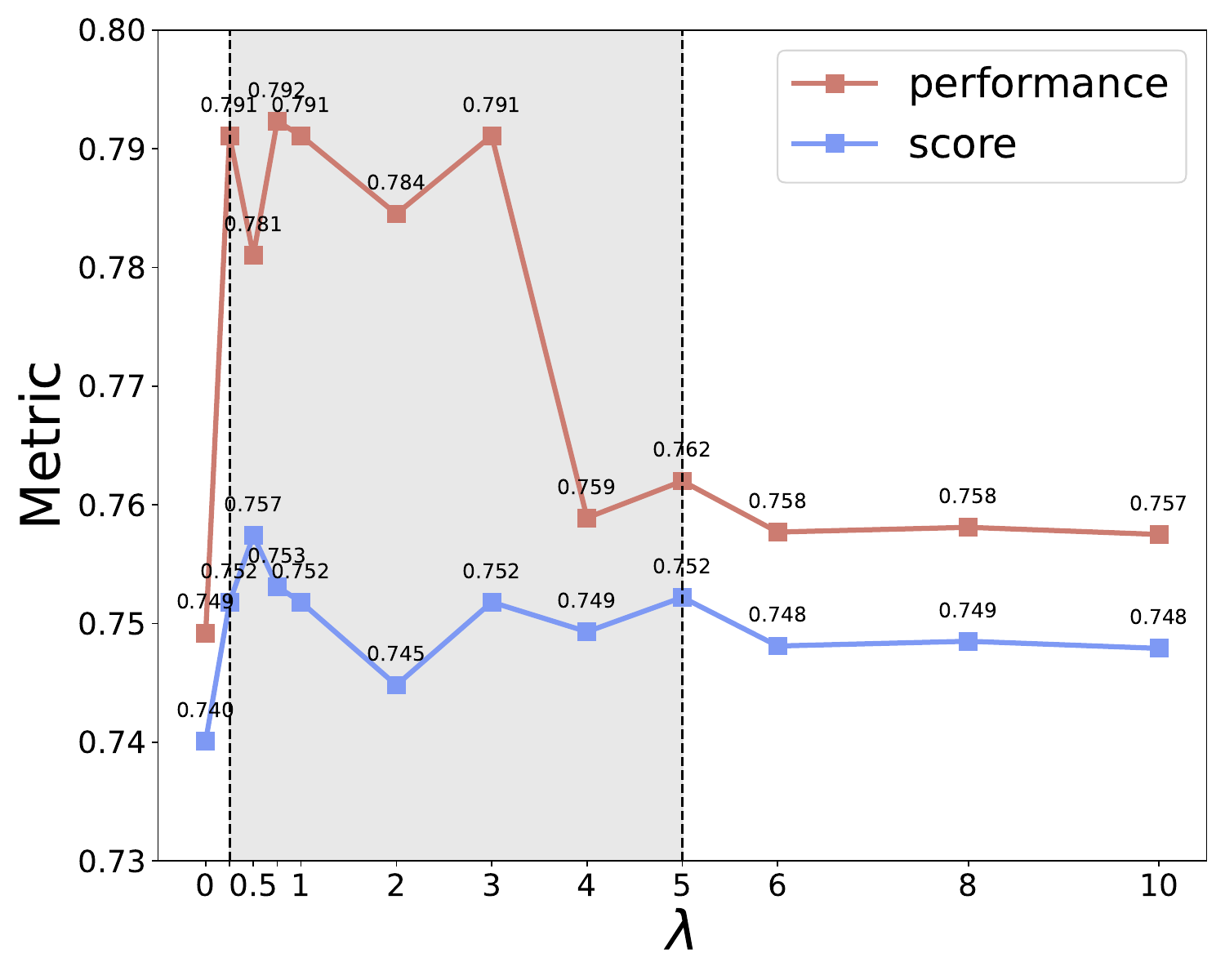}
  \caption{Effects of $\lambda$ in the \textit{Balance} scenario.}
  \label{fig:effects_of_lambda}
  \vspace{-0.5cm}
\end{figure}

\section{Detailed Results of Effects of $\lambda$ on Optimization}
\label{sec:full_results_of_effects_of_lambda}

Tab.\ref{tab:effects_of_lambda} shows the effects of the contrastive loss weight $\lambda$ on the optimization of RadialRouter. Experiments are conducted in the \textit{Balance} scenario, visualized in Fig.\ref{fig:effects_of_lambda}. As can be seen, the highest score is achieved when $\lambda=0.5$ and RadialRouter demonstrates insensitivity across a wide range of $\lambda\in[0.25,5]$, which confirms the robustness of our method and provides greater flexibility for the selection of $\lambda$ in practice.

\end{document}